\documentclass{article}

\usepackage[T1]{fontenc}
\usepackage{microtype}
\usepackage{graphicx}
\graphicspath{ {./images/} }
\usepackage{natbib}

\usepackage{booktabs}
\usepackage{multirow}
\usepackage[table]{xcolor}
\usepackage{colortbl}
\usepackage{amsfonts}
\usepackage{nicefrac}
\usepackage{listings}
\usepackage[linesnumbered,ruled,vlined]{algorithm2e}
\usepackage{enumitem}
\usepackage{csquotes}
\usepackage{lipsum}
\usepackage{arxiv}

\usepackage{hyperref}

\lstdefinestyle{mypython}{
  language=Python,
  backgroundcolor=\color{gray!10},
  basicstyle=\ttfamily\small,
  keywordstyle=\color{blue}\bfseries,
  commentstyle=\color{green!50!black}\itshape,
  stringstyle=\color{red},
  showstringspaces=false,
  frame=single,
  numbers=left,
  numberstyle=\tiny\color{gray},
  stepnumber=1,
  breaklines=true,
}

\usepackage{xcolor,colortbl,pgf,siunitx}
\usepackage{xcolor}

\definecolor{effectPos}{HTML}{2E8B57}   
\definecolor{effectNeg}{HTML}{B22222}   
\definecolor{effectZero}{gray}{0.95}    

\newcommand{\effectcolor}[1]{%
  \begingroup
    \pgfmathsetmacro{\val}{#1}%
    \pgfmathsetmacro{\aval}{abs(\val)}%
    \pgfmathsetmacro{\tint}{40 + min(55, 55 * \aval * 3)}%
    \ifdim \val pt > 0pt
      \edef\col{effectPos!\tint!white}%
    \else
      \edef\col{effectNeg!\tint!white}%
    \fi
    \pgfmathsetmacro{\isDark}{\tint < 65 ? 1 : 0}%
    \ifnum\isDark=1
      \colorbox{\col}{\color{white}\strut\num[round-mode=places,round-precision=2]{\val}}%
    \else
      \colorbox{\col}{\color{black}\strut\num[round-mode=places,round-precision=2]{\val}}%
    \fi
  \endgroup
}

\definecolor{lightgreen}{rgb}{0.88, 1, 0.88}
\definecolor{lightgray}{gray}{0.95}
\definecolor{blue1}{RGB}{222,235,247}
\definecolor{blue2}{RGB}{198,219,239}
\definecolor{blue3}{RGB}{158,202,225}
\definecolor{blue4}{RGB}{107,174,214}
\definecolor{red1}{RGB}{254,224,210}
\definecolor{red2}{RGB}{252,187,161}
\definecolor{red3}{RGB}{252,146,114}
\definecolor{red4}{RGB}{222,45,38}

\title{MindSET: Advancing Mental Health Benchmarking through Large-Scale Social Media Data}

\author{
Saad Mankarious$^{1}$ \quad
Ayah Zirikly$^{1}$ \quad
Daniel Wiechmann$^{2,3}$ \quad
Elma Kerz$^{3}$ \quad
Edward Kempa$^{4}$ \quad
Yu Qiao$^{3}$ \\[4pt]
$^{1}$School of Engineering and Applied Science, George Washington University, Washington, D.C. 20037 \\
$^{2}$Institute for Logic, Language \& Computation, University of Amsterdam, the Netherlands \\
$^{3}$Exaia Technologies, Germany \\
$^{4}$Department of Computer and Information Science and Engineering, University of Florida, USA \\[4pt]
\texttt{\{saadm, ayah.zirikly\}@gwu.edu} \\
\texttt{d.wiechmann@uva.nl, kempaedward@ufl.edu} \\
\texttt{\{e.kerz, y.qiao\}@exaia-tech.com}
}

\begin{document}

\maketitle

\begin{center}
\vspace{-0.5em}
\textbf{GitHub:} \href{https://github.com/fibonacci-2/mindset}{\texttt{github.com/fibonacci-2/mindset}} \quad
\raisebox{-0.2em}{\includegraphics[height=1em]{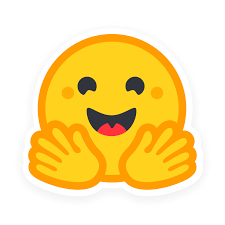}}\ 
\href{https://huggingface.co/datasets/smankarious/mindset}{\texttt{huggingface.co/datasets/smankarious/mindset}}
\vspace{1em}
\end{center}

\begin{abstract}
Social media data has become a vital resource for studying mental health, offering real-time insights into thoughts, emotions, and behaviors that traditional methods often miss. Progress in this area has been facilitated by benchmark datasets for mental health analysis; however, most existing benchmarks have become outdated due to limited data availability, inadequate cleaning, and the inherently diverse nature of social media content (e.g., multilingual and harmful material). We present a new benchmark dataset, \textbf{MindSET}, curated from Reddit using self-reported diagnoses to address these limitations. The annotated dataset contains over \textbf{13M} annotated posts across seven mental health conditions, more than twice the size of previous benchmarks. To ensure data quality, we applied rigorous preprocessing steps, including language filtering, and removal of Not Safe for Work (NSFW) and duplicate content. We further performed a linguistic analysis using LIWC to examine psychological term frequencies across the eight groups represented in the dataset. To demonstrate the dataset’s utility, we conducted binary classification experiments for diagnosis detection using both fine-tuned language models and Bag-of-Words (BoW) features. Models trained on MindSET consistently outperformed those trained on previous benchmarks, achieving up to an \textbf{18-point} improvement in F1 for Autism detection. Overall, MindSET provides a robust foundation for researchers exploring the intersection of social media and mental health, supporting both early risk detection and deeper analysis of emerging psychological trends.
\end{abstract}

\section{Background}
\begin{figure*}[h!]
    \centering
    \includegraphics[width=.7\linewidth]{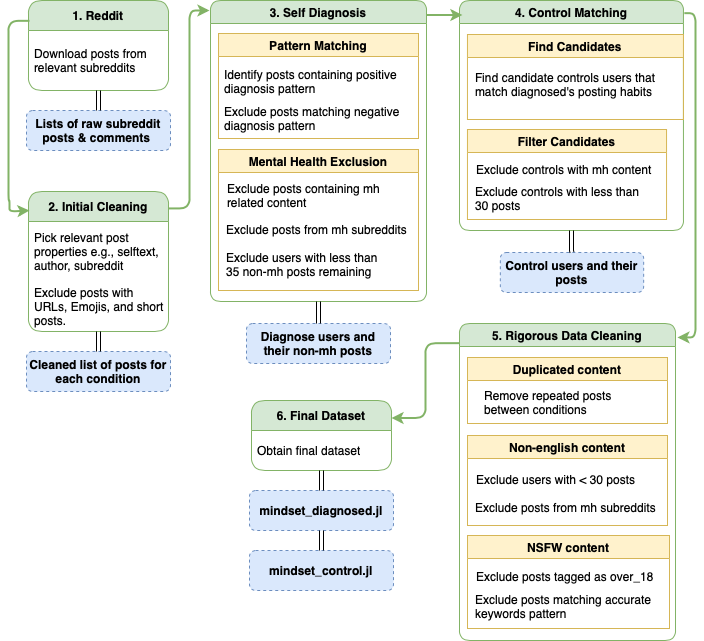}
    \caption{Step-by-step end-to-end pipeline for dataset construction. The output of each stage is shown in the blue rectangles below.}
    \label{fig:pipeline}
\end{figure*}

Traditionally, mental health diagnoses have relied on clinical interviews, self-report questionnaires, and standardized assessments. While these methods are effective, they are often time-consuming and may fail to capture real-time fluctuations in an individual's mental state. In contrast, user-generated content on social media platforms offers a valuable alternative, reflecting users’ thoughts, emotions, and behaviors as they naturally occur~\cite{reece2017instagram, park2013}. Consequently, linguistic analysis of social media posts to identify patterns indicative of mental health conditions has become an increasingly prominent area of research~\cite{reece2017instagram}. Such approaches enable continuous monitoring and early intervention, potentially reaching individuals who might not engage with traditional mental health services~\cite{park2013}.


Effective mental health analysis using social media data requires comprehensive, annotated datasets that capture diverse conditions and ensure sufficient sample sizes to enhance the reliability of findings. Historically, datasets such as the Social Media and Mental Health Dataset (SMHD), which includes data for nine mental health conditions~\cite{cohan2018smhd}, provided a strong benchmark for research due to their scale and methodological rigor. Using SMHD, \cite{dinu2021automatic} achieved state-of-the-art performance with an F1 score of 81 for diagnosing eating disorders. Other notable datasets include the CLPsych 2015 Shared Task dataset, which focused on detecting depression and PTSD from Twitter data~\cite{coppersmith2015adhd}; the Reddit Self-Reported Depression Diagnosis (RSDD) dataset, designed for identifying depression based on self-reported diagnoses~\cite{yates2017depression}; and the eRisk dataset, which supports early detection of mental illness through longitudinal user interactions~\cite{losada2016erisk}.

Despite the availability of prior resources, two significant limitations remain. First, these datasets relied on \texttt{Pushshift}, the official Reddit API, which is currently discontinued by Reddit, preventing sharing and use of such datasets.\footnote{We confirmed this limitation through correspondence with the original authors of SMHD.}  

Second, social media content on Reddit often contains substantial noise, including repeated submissions (users reposting the same content to attract more responses), Not Safe for Work (NSFW) material (e.g., pornographic or sexually explicit content, which is particularly common due to Reddit’s anonymity), and multilingual posts that conflict with English-focused analyses.\footnote{Our goal is to build a dataset exclusively focused on English content, as opposed to a multilingual resource.} Without rigorous preprocessing, models risk learning spurious correlations rather than meaningful indicators of mental health conditions \cite{olteanu2019social, chancellor2020methods}. Moreover, such harmful content poses ethical and safety concerns for researchers handling the data, study participants, and end users of downstream systems, as highlighted in prior work \cite{kirk2022handling}. Accordingly, we apply strict filtering to exclude this type of content. To the best of our knowledge, prior datasets have not systematically addressed these challenges \cite{chancellor2020methods, yates2017depression, yates2017rsdd, coppersmith2015adhd}.

We introduce \textbf{MindSET}, a large-scale and rigorously cleaned dataset designed to advance research in social media–based mental health analysis. MindSET fills key gaps in existing resources by offering a dataset nearly twice the size of prior benchmarks, while ensuring high data quality and integrity. Our main contributions are summarized as follows:

\begin{itemize}
    \item We compile a large-scale Reddit dataset for mental health research, exceeding the size of existing benchmarks by more than twofold.
    \item We apply comprehensive preprocessing steps, including language filtering, duplicate detection, and NSFW content removal, to ensure data reliability and reduce noise.
    \item We establish new state-of-the-art benchmarks for social media–driven mental health analysis through the combination of high-quality data and improved model performance.
    \vspace{-8pt}
\end{itemize}


\begin{table*}[h!]
\centering
\small
\begin{tabular}{ccccc}
\hline
\\[1pt]
\textbf{Condition} & \textbf{SMHD, Cohan} & \textbf{SMHD, Dinu} & \textbf{MindSET, BERT} & \textbf{MindSET, XGBoosts} \\\\[1pt]
\hline
Depression  & 53 & 70 & 76\textbf{ (+6)} & 86\textbf{ (+16)} \\
OCD        & 44 & 75 & 77\textbf{ (+2)} & 85\textbf{ (+10)} \\
Bipolar    & 57 & 75 & 78\textbf{ (+3)} & 89\textbf{ (+14)} \\
ADHD       & 47 & 71 & 75\textbf{ (+4)} & 85\textbf{ (+14)} \\
PTSD       & 57 & 76 & 80\textbf{ (+4)} & 89\textbf{ (+13)} \\
Autism      & 49 & 71 & 77\textbf{ (+6)} & 89\textbf{ (+18)} \\
Anxiety     & 54 & 73 & 77\textbf{ (+4)} & 87\textbf{ (+14)} \\
\hline
\end{tabular}
\caption{Binary Classification Performance from previous baselines \cite{cohan2018smhd, dinu2021automatic}, and ours. Our models consistently outperform both baselines across \emph{all} conditions, with biggest improvement on Autism detection by \textbf{+18} F1 points.}
\label{tab:baseline-comparason}
\end{table*}

\section{Related Work}

Recent research has increasingly leveraged social media data for the early diagnosis and analysis of mental health conditions. \cite{coppersmith2015adhd} demonstrated the effectiveness of using Twitter data for detecting depression and PTSD, highlighting social media platforms as valuable resources for mental health research. Their dataset comprised approximately 1,800 Twitter users with self-identified diagnoses of depression and PTSD, collected through user statements and interactions. The SMHD (Shared Mental Health Dataset) further advanced this line of work by providing labeled social media posts for multiple mental health conditions, establishing a strong foundation for training machine learning models to detect disorders such as depression and anxiety~\cite{cohan2018smhd}. Expanding on crisis-related research, CLPsych introduced annotated Reddit posts from shared tasks focused on suicide prevention and crisis identification, offering a critical resource for studying urgent mental health scenarios~\cite{milne2016clpsych}. Similarly, the RSDD (Reddit Self-Reported Depression Diagnosis) dataset specifically targets depression detection, utilizing self-reported diagnosis posts to enable more accurate predictive modeling of depressive behaviors~\cite{yates2017rsdd}. In parallel, the eRisk initiative proposed datasets for early risk detection, structuring challenges that evaluate algorithmic performance over time in predicting mental health deterioration~\cite{losada2016erisk}.

This approach automatically detects users who self-disclose a mental health diagnosis by scanning their text for specific diagnostic terms and patterns. First established by \cite{coppersmith2015adhd} and later enhanced by researchers like \cite{yates2017depression} and \cite{cohan2018smhd}, this method enabled the creation of large-scale, annotated datasets from platforms like Reddit without manual effort, solving key scalability issues. Its utility extends beyond English to other languages such as German and Arabic \cite{zanwar-etal-2023-smhd, mankarious2025carma}. This proven adaptability underscores the method's robustness and motivates its application to underrepresented languages like Arabic.

Several benchmark datasets have been introduced to support mental health diagnosis research using social media data. The SMHD dataset, released in 2017 by \cite{cohan2018smhd}, provided a comprehensive resource constructed from self-reported diagnoses across multiple conditions. Subsequent work by \cite{dinu2021automatic} established stronger baselines, achieving higher accuracy scores using a pretrained BERT model trained on a smaller subset of SMHD data. Furthermore, \cite{shing-etal-2018-expert} used self-reported diagnosis to establish a baseline rubric for assessing suicide risk through crowd sourcing and expert annotations of Reddit data.

However, since most of these datasets were built using the Pushshift API, which has since been restricted by Reddit, many of these resources are now obsolete and unavailable for reuse. To overcome this limitation, we constructed our dataset directly from Reddit using an alternative API that complies fully with Reddit’s current privacy policies. We adopt the self-reported diagnosis patterns introduced by \cite{coppersmith2018smhd}. Furthermore, we implement rigorous data-cleaning procedures, including non-English content filtering, duplicate removal, and NSFW (Not Safe For Work) content exclusion, to minimize bias and enhance dataset reliability.

\vspace{-8pt}
\section{Dataset Construction}

Using the Arctic Shift API\footnote{\url{https://github.com/ArthurHeitmann/arctic_shift}}, we collect Reddit contributions made between January 2018 and September 2024 to a list of mental health related subreddits. We then apply a modified version of the self-diagnosis pattern introduced by \cite{coppersmith2018smhd} and later enhanced by \cite{yates2017depression} and \cite{cohan2018smhd} to identify users who explicitly self-report a diagnosis of a mental health condition (Subsection~\ref{subsec:self_diagnosis}). Each self-diagnosed user is then matched with a set of control users exhibiting similar posting habits (Subsection~\ref{subsec:control_matching}). Finally, we apply a three-step cleaning procedure to mitigate potential sources of bias and enhance dataset integrity (Sections \ref{cleaning-section}).
\vspace{-8pt}

\subsection{Identification of Diagnosed Users}
\label{subsec:self_diagnosis}
We employ a high-precision self-diagnosis pattern to identify users with mental health conditions. The patterns consist of two components: one that matches a self-reported diagnosis (e.g., “diagnosed with”), and another that maps relevant keywords to the 7 mental health disorders (e.g., ADHD and Anxiety). As shown in Step~3 of Figure~\ref{fig:pipeline}, we detect users who explicitly state that they have been diagnosed by matching their posts against patterns similar to Statement 1 in Figure~\ref{fig:diagnosis-quote}. It is important to distinguish this from \emph{tentative} or \emph{self-suspected} cases, where users merely express uncertainty about their condition (as in Statement~2 of Figure~\ref{fig:diagnosis-quote}), rather than reporting an established diagnosis. The pattern is designed to capture self-diagnosis statements with an accuracy of 96.4\% according to \citealt{coppersmith2018smhd}\footnote{The pattern also accounts for negative diagnosis, where users say \textit{"I was \emph{not} diagnosed with"}. False positive cases are instance when the user is referring to or quoting someone else, e.g., \textit{"my brother} was diagnosed".}. In absence of clinical ground truth, we treat this as gold labels, following previous established literature \cite{shing-etal-2018-expert, cohan2018smhd}.

\begin{figure}[ht]
  \centering
  \begin{minipage}{0.5\textwidth}
    \centering
    \itshape
    \rule{\linewidth}{0.1pt}

    Statement 1: \textbf{"I was officially diagnosed with depression by my doctor"} \\[6pt]
    Statement 2: \textbf{"I think I might have depression"}

    \rule{\linewidth}{0.1pt}
\vspace{-8pt}
\caption{Distinction between confirmed self-reported diagnoses (Statement~1) and tentative or uncertain expressions (Statement~2). Only patterns resembling Statement~1 were used.}
    \vspace{-8pt}

    \label{fig:diagnosis-quote}
  \end{minipage}
\end{figure}

A key step in \cite{cohan2018smhd}'s implementation of self-reported diagnosis, which we thereafter adopted, is the exclusion of mental health content from the dataset. This is to prevent making the task of identifying mental health conditions trivial (e.g., by accidentally including a data document where users say \textit{"life is very depressing"}). We expanded the exclusion pattern by updating the list of subreddits considered as Mental Health forum. For example, adding \texttt{r/AnxietyRecovery} and \texttt{r/schizoaffective}, with 133K and 110K members respectively, were included to ensure that the dataset minimizes explicit mental health discussions. This same list is also applied when identifying potential control users (see Section~\ref{subsec:control_matching}).
\vspace{-8pt}

\subsection{Control Matching}
\label{subsec:control_matching}

Prior research has emphasized the importance of constructing control groups that closely match the posting habits and frequency of diagnosed users to enable balanced and unbiased discriminative analysis~\cite{cohan2018smhd, dinu2021automatic, coppersmith2015adhd}. Following this approach, we construct a large control group designed to mirror the behavioral characteristics of the diagnosed cohort.

To ensure comparable posting activity, the number of posts made by each control user is constrained to fall within a defined range relative to their matched diagnosed user, consistent with the methodology described in~\cite{coppersmith2018smhd}. Additionally, we exclude and remove all users who post any mental health–related content using a keyword-based detection pattern. This step minimizes the likelihood of mistakenly including diagnosed individuals in the control group, similar to our approach in excluding MH content from the diagnosed group (Section \ref{subsec:self_diagnosis}).

The previous two factors pose a significant challenge obtaining control users as they diminish the control group significantly \cite{cohan2018smhd}. To combat this, we kept the previous two conditions but relaxed the limit on the number of posts from 50 to 35 compared to SMHD, which enabled us to around 4-5 control users per diagnosed user compared to 9 users SMHD obtained.

\section{Dataset Cleaning}
\label{cleaning-section}
\begin{figure*}[h!]
    \centering
    \includegraphics[width=\linewidth]{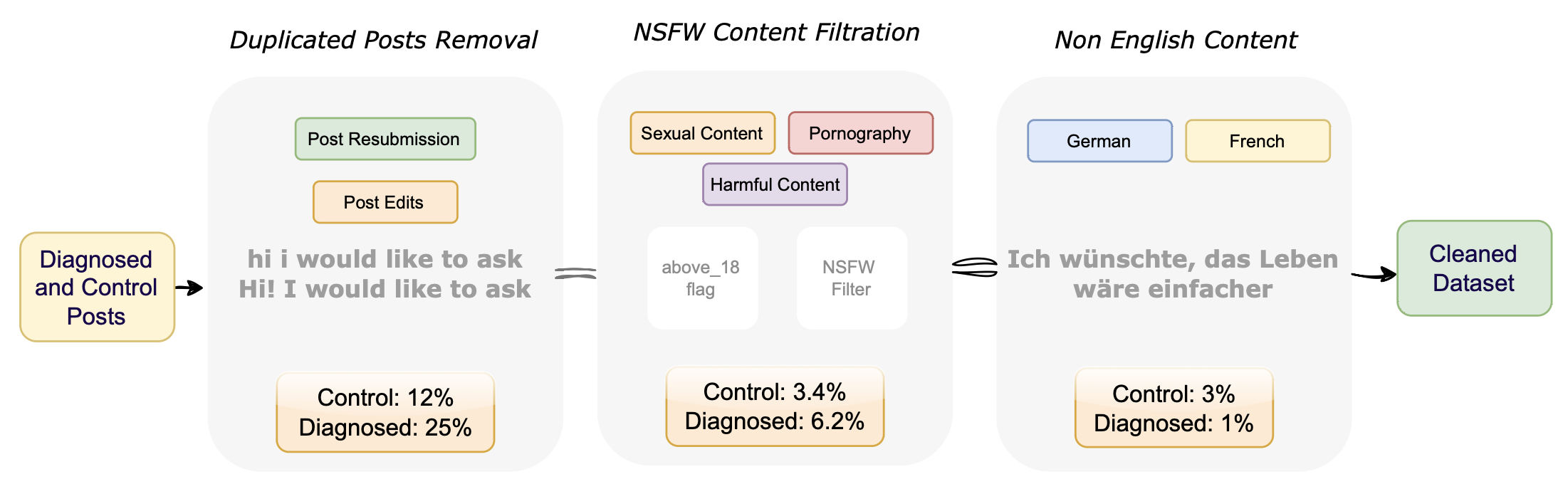}
\caption{Cleaning pipeline. Each step shows the percentage decrease in data volume. Example content filtered at each stage is displayed, except for the middle step, which contains inappropriate material.}
    \label{fig:cleaning-pipeline}
\end{figure*}

We implemented a rigorous data cleaning pipeline (Figure~\ref{fig:cleaning-pipeline}) to filter out harmful or low-quality content that may introduce noise and obscure meaningful mental health signals. Figure \ref{fig:cleaning-pipeline} shows examples of such content. Although such datasets are derived from open discussions, where user-generated content is inherently difficult to regulate, we aimed to enhance both data integrity and ethical standards through systematic filtering. Figure~\ref{fig:cleaning-pipeline} outlines our cleaning process, which is described in detail in the following subsections.

\subsection{Non-English Content}
Given the multilingual nature of Reddit, it was essential to assess the extent of non-English content across the dataset. We employed the \texttt{langdetect} Python library\footnote{\url{https://pypi.org/project/langdetect/}} to identify non-English posts for each condition. As shown in The proportion of non-English content was relatively low among diagnosed users but higher among controls (3.02\%). This difference likely reflects the broader and more diverse nature of discussions within the control group, compared to the more focused discourse of diagnosed users.

Ensuring linguistic uniformity is crucial, as even a small proportion of non-English text can introduce noise, diminishing the effectiveness of natural language processing tasks such as sentiment analysis and topic modeling~\cite{baldwin2010language}. Consequently, all non-English content was removed from the final dataset.


\vspace{-8pt}
\subsection{Duplicated Posts}
During preprocessing, we identified a substantial proportion of duplicated posts arising from two main causes: (1) Reddit users frequently resubmit identical posts to enhance visibility, and (2) control users are often matched to multiple diagnosed users across conditions, leading to repeated instances of the same post.


Eliminating duplicates is crucial to ensure data validity and analytic reliability. Retaining them could artificially inflate certain linguistic patterns, biasing models toward overrepresented content and diminishing generalizability \cite{lee2021deduplication}. Consequently, all duplicated content was removed from the final dataset to uphold data integrity and robustness.

\subsection{NSFW Content}
The inclusion of NSFW (Not Safe For Work) material in mental health datasets poses serious ethical and methodological challenges. Such content introduces noise. For instance, NSFW content was significantly more prevalent in the control group than the diagnosed groups in our dataset which could bias model learning toward irrelevant features in addition to posing a potential risk towards researchers and research participants who are using the dataset \cite{kirk2022handling}. To ensure the dataset’s ethical soundness and analytical rigor, we employed a two-stage filtering strategy to identify and exclude NSFW content.

\begin{itemize}
\vspace{-8pt}
    \item \textbf{Reddit API flag:} We used the \texttt{over\_18} attribute provided by the Reddit API to automatically exclude all posts flagged as over 18, along with all comments associated with such posts.
    \vspace{-8pt}
    \item \textbf{Pattern-based filtering:} We constructed an automatic keyword filter comprising 238 terms, primarily derived from the control dataset, that are frequently associated with explicit or adult content.
    \vspace{-4pt}
\end{itemize}

Applying \texttt{over\_18} flag identified \textbf{3.45\%} of posts from diagnosed users and \textbf{2.70\%} from control users as NSFW. Following this, the keyword-based filter was applied, matching each post against the sensitive term list. Posts with one or more matches were excluded from the final dataset, corresponding to the highest proportion of NSFW content. Combining both detection methods reduced the final dataset by \textbf{9.68\%} and \textbf{6.14\%} for the diagnosed and control groups, respectively. Although this reduction is substantial, it was essential to maintain high data quality. Even after filtering, our dataset remained more than twice as large as previous benchmarks, underscoring its robustness and scalability.



\begin{table*}[ht]
\centering
\label{tab:summary}
\begin{tabular}{lrrrrrrr}
\hline
\textbf{Condition} & \textbf{Users} & \textbf{Texts} & \textbf{Sentences} & \textbf{Words} & \textbf{Texts/User} & \textbf{Sentences/Text} & \textbf{Words/Text} \\
\hline
Control    & 170{,}611 & 31.9M & 131.9M & 1.6B  & 187.23 & 4.13 & 51.46 \\
ADHD       & 18{,}669  & 4.9M  & 26.1M  & 381.4M & 260.23 & 5.37 & 78.50 \\
Autism     & 9{,}008   & 2.4M  & 12.6M  & 186.9M & 262.75 & 5.32 & 78.96 \\
Bipolar    & 7{,}162   & 1.7M  & 9.3M   & 130.9M & 236.12 & 5.48 & 77.38 \\
Depression & 6{,}530   & 1.5M  & 8.2M   & 117.8M & 236.50 & 5.32 & 76.30 \\
Anxiety    & 3{,}772   & 1.0M  & 5.7M   & 81.5M  & 278.23 & 5.42 & 77.64 \\
OCD        & 2{,}412   & 639K  & 3.2M   & 47.8M  & 265.12 & 5.08 & 74.74 \\
PTSD       & 2{,}391   & 616K  & 3.4M   & 48.6M  & 257.72 & 5.54 & 78.95 \\
\hline
\end{tabular}
\caption{Post-level statistics of the final cleaned dataset.}

\end{table*}

\vspace{-8pt}
\section{Dataset Exploration}
We analyze the collected dataset to assess its scale and lexical characteristics across both self-diagnosed and control groups. Section~\ref{descriptive-stats} presents descriptive statistics of the cleaned dataset, while Section~\ref{liwc} details our psychological term frequency analysis.
\vspace{-8pt}
\subsection{Descriptive Statistics}
\label{descriptive-stats}

Table~\ref{tab:summary} summarizes the statistics of the final dataset across seven diagnosed conditions and the control group after all three cleaning stages. As shown, the most prevalent conditions in the dataset are \textit{ADHD}, \textit{Autism}, and \textit{Bipolar}.

The \textit{ADHD} and \textit{Autism} groups exhibit higher average text and word counts per user compared to other conditions, suggesting a greater degree of engagement and textual output. While the average number of sentences per text remains relatively consistent across conditions, variations in average word counts per text may reflect differences in linguistic style, expression length, or cognitive processing patterns.
\vspace{-8pt}

\subsection{Psychological Term Category Analysis}
\label{liwc}
\begin{table*}[h]
\centering
\renewcommand{\arraystretch}{0.7} 
\small 

\begin{tabular}{lcccccccc}
\hline
\textbf{LIWC Feature} & \textbf{ADHD} & \textbf{Anxiety} & \textbf{Autism} & \textbf{Bipolar} & \textbf{Depression} & \textbf{OCD} & \textbf{PTSD} \\
\hline
Affect & \effectcolor{0.16} & \effectcolor{0.34} & \effectcolor{0.27} & \effectcolor{0.37} & \effectcolor{0.40} & \effectcolor{0.38} & \effectcolor{0.48} \\

Analytic & \effectcolor{-0.63} & \effectcolor{-0.99} & \effectcolor{-0.83} & \effectcolor{-0.87} & \effectcolor{-0.97} & \effectcolor{-1.06} & \effectcolor{-1.08} \\

Authentic & \effectcolor{0.38} & \effectcolor{0.59} & \effectcolor{0.40} & \effectcolor{0.51} & \effectcolor{0.47} & \effectcolor{0.56} & \effectcolor{0.52} \\
Cognition & \effectcolor{0.52} & \effectcolor{0.54} & \effectcolor{0.72} & \effectcolor{0.44} & \effectcolor{0.58} & \effectcolor{0.69} & \effectcolor{0.55} \\
Emotion & \effectcolor{0.47} & \effectcolor{0.82} & \effectcolor{0.62} & \effectcolor{0.78} & \effectcolor{0.76} & \effectcolor{0.85} & \effectcolor{0.93} \\
Tone & \effectcolor{0.25} & \effectcolor{0.29} & \effectcolor{-0.77} & \effectcolor{0.23} & \effectcolor{0.20} & \effectcolor{0.30} & \effectcolor{0.35} \\

Past & \effectcolor{-0.05} & \effectcolor{0.04} & \effectcolor{0.42} & \effectcolor{0.03} & \effectcolor{0.02} & \effectcolor{0.05} & \effectcolor{0.01} \\
Present & \effectcolor{0.10} & \effectcolor{0.12} & \effectcolor{0.33} & \effectcolor{0.08} & \effectcolor{0.09} & \effectcolor{0.15} & \effectcolor{0.11} \\
Future & \effectcolor{-0.02} & \effectcolor{0.01} & \effectcolor{0.08} & \effectcolor{0.00} & \effectcolor{-0.01} & \effectcolor{0.03} & \effectcolor{0.02} \\

Social & \effectcolor{0.15} & \effectcolor{0.18} & \effectcolor{0.60} & \effectcolor{0.20} & \effectcolor{0.21} & \effectcolor{0.25} & \effectcolor{0.22} \\
Friends & \effectcolor{0.10} & \effectcolor{0.09} & \effectcolor{0.31} & \effectcolor{0.08} & \effectcolor{0.10} & \effectcolor{0.12} & \effectcolor{0.14} \\
Confidence & \effectcolor{0.02} & \effectcolor{0.03} & \effectcolor{0.30} & \effectcolor{0.01} & \effectcolor{0.00} & \effectcolor{0.04} & \effectcolor{0.03} \\

Power & \effectcolor{-0.07} & \effectcolor{-0.06} & \effectcolor{0.09} & \effectcolor{-0.08} & \effectcolor{-0.09} & \effectcolor{-0.07} & \effectcolor{-0.05} \\
Risk & \effectcolor{0.20} & \effectcolor{0.22} & \effectcolor{0.51} & \effectcolor{0.21} & \effectcolor{0.19} & \effectcolor{0.23} & \effectcolor{0.24} \\
Anger & \effectcolor{-0.03} & \effectcolor{-0.01} & \effectcolor{0.75} & \effectcolor{-0.04} & \effectcolor{-0.02} & \effectcolor{-0.10} & \effectcolor{-0.05} \\
Sadness & \effectcolor{0.08} & \effectcolor{0.11} & \effectcolor{0.59} & \effectcolor{-0.12} & \effectcolor{0.10} & \effectcolor{0.15} & \effectcolor{0.14} \\
Anxiety & \effectcolor{0.13} & \effectcolor{0.19} & \effectcolor{0.78} & \effectcolor{0.17} & \effectcolor{0.20} & \effectcolor{0.18} & \effectcolor{0.21} \\
\hline
\end{tabular}

\caption{Effect sizes (Cohen’s d) of LIWC features comparing diagnosed and control groups. Blue shading indicates features more prevalent in the diagnosed group, and red shading indicates features more prevalent in the control group.}
\label{tab:liwc_effect_sizes}
\end{table*}

LIWC \cite{pennebaker2003psychological} is a widely used psycholinguistic tool that extracts category-specific lexical features with psychological relevance. These categories capture both linguistic style and psychological dimensions of language use (e.g., cognitive and affective attributes). For each user, we compute LIWC category scores from their posts and compare these across diagnosed and control users using Welch’s \texttt{t-test} \cite{welch1947generalization}, as per \cite{cohan2018smhd}. To control for multiple comparisons, we apply the Bonferroni correction to adjust p-values, and report \texttt{Cohen’s d} effect sizes \cite{cohen2013statistical} to quantify the magnitude of group differences. Table~\ref{tab:liwc_effect_sizes} presents per-condition term categories and their corresponding effect sizes. Larger effect sizes indicate stronger discriminative power between diagnosed and control users.

\textbf{Overall Insights.} Several LIWC categories exhibit varying effect sizes (\emph{Cohen's d}) distinguishing diagnosed and control groups. Control users consistently show higher usage of \texttt{Analytic} language, aligning with findings from \cite{cohan2018smhd}, where the \texttt{Numbers} category also showed a large effect size. Conversely, all diagnosed groups demonstrate a consistent increase in \texttt{Authentic} language, reflecting more personal, honest, and self-disclosing expression compared to controls, a pattern supported by prior research \cite{bucci1981language, pennebaker2015liwc, watkins2002rumination, van2014web}. 

Furthermore, categories related to social connectedness and confidence, such as \texttt{Social}, \texttt{Friends}, and \texttt{Confidence}, are more prevalent among diagnosed users, including those matched against Anxiety and Depression groups. This aligns with earlier work by \cite{murphy1991depression} and corroborates similar findings reported in \cite{cohan2018smhd}.

Next, we examine each diagnosed group in the dataset in greater detail:

\vspace{-4pt}
\textbf{Autism.}
The Autism group exhibits the most distinctive overall profile, with pronounced differences in social language, temporal orientation, and specific emotional expressions relative to other conditions, which is consistent with existing literature that showed the diverse social and emotional challenges of Autistic people on Reddit \cite{fong2025autism}. Large effect sizes are observed for decreased \texttt{Analytic} thinking ($d = -0.83$) and increased \texttt{Cognition} ($d = 0.72$), \texttt{Social} references ($d = 0.60$) and emotional features including \texttt{Anger} ($d = 0.75$), \texttt{Sadness} ($d = 0.59$), and notably \texttt{Anxiety} ($d = 0.78$). This group also displays substantially lower \texttt{Tone} ($d = -0.77$) and increased focus on \texttt{Past} ($d = 0.42$) and \texttt{Present} ($d = 0.33$) time frames.

\textbf{ADHD.}
The ADHD group exhibits a moderate decrease in \texttt{Analytic} thinking ($d = -0.63$) and moderate increases in \texttt{Authentic} ($d = 0.38$) and \texttt{Cognition} ($d = 0.52$) features. Other effects are generally small, with minimal differences in temporal focus, social domains, and emotional expression compared to controls.

\textbf{Anxiety.}
The Anxiety group shows a large decrease in \texttt{Analytic thinking} ($d = -0.99$) alongside large increases in \texttt{Emotion} ($d = 0.82$) and moderate increases in \texttt{Authentic} ($d = 0.59$), \texttt{Affect ($d = 0.34$)}, and \texttt{Cognition} ($d = 0.54$). Effects within temporal and social domains remain small, though slight increases are observed across most emotional categories.

\textbf{Depression.}
The Depression group exhibits a pattern similar to Anxiety, characterized by a large decrease in \texttt{Analytic} thinking ($d = -0.97$) and increases in \texttt{Emotion} ($d = 0.76$), \texttt{Cognition} ($d = 0.58$), and \texttt{Authentic} ($d = 0.47$). Moderate effects are found for \texttt{Affect} ($d = 0.40$) and social domains, with minimal variation in temporal focus. The similarity between Depression and Anxiety profiles is consistent with prior findings in the literature, as \cite{cohan2018smhd, murphy1991depression} observed.

\textbf{Bipolar.}
The Bipolar group shows a large decrease in \texttt{Analytic} thinking ($d = -0.87$) accompanied by increases in \texttt{Emotion} ($d = 0.78$), \texttt{Authentic} ($d = 0.51$), and \texttt{Affect} ($d = 0.37$). A moderate increase in \texttt{Cognition} ($d = 0.44$) is also observed, while temporal and social domains show minimal differences from controls.

\vspace{-8pt}
\section{Experiments}
To demonstrate how the mental health signal embedded in the dataset can be effectively utilized, we conduct two classification experiments using pretrained models finetuning and shallow classifiers trained on Bag of Word features. Both experiments are balanced binary classification where we sample same number of samples from diagnosed class and the control class. 

\vspace{-8pt}
\subsection{Bag of Words Features (BoW)}

\begin{table}[h]
\centering
\small
\begin{tabular}{cc c c c c}
\hline
\textbf{Condition} & \textbf{Model} & \textbf{Acc.} & \textbf{Prec.} & \textbf{Rec.} & \textbf{F1} \\
\hline
ADHD       &  SVM  & 86 & 86 & 84 & 85 \\
Depression &  SVM  & 86 & 88 & 83 & 86 \\
Anxiety    &  SVM  & 87 & 87 & 87 & 87 \\
Bipolar    & XGBoost     & 89 & 90 & 87 & 89 \\
OCD        & XGBoost     & 86 & 89 & 82 & 85 \\
PTSD       & XGBoost     & 90 & 91 & 87 & 89 \\
Autism     & XGBoost     & 89 & 91 & 87 & 89 \\
\hline
\end{tabular}
\caption{Best model performance per condition using TF-IDF weighted Bag-of-Words features.}
\label{tab:finetuning-on-bow}
\end{table}

\label{bow-section}
We motivated training three classifiers, \texttt{SVM}, \texttt{XGBoost}, and \texttt{Logistic Regression}, using BoW features as per \cite{cohan2018smhd}. User posts were aggregated, tokenized into individual words, lowercased, and filtered to remove infrequent terms. The remaining tokens were statistically weighted and normalized to emphasize the most distinctive features. Our results are shown in Table \ref{tab:finetuning-on-bow}.
\vspace{-8pt}
\subsection{BERT Finetuning}
\label{finetuning}

\begin{table}[h]
\centering
\small
\begin{tabular}{l c c c c}
\hline
\textbf{Label} & \textbf{Acc.} & \textbf{Prec.} & \textbf{Rec.} & \textbf{F1} \\
\hline
PTSD       & 86 | 81 & 83 | 80 & 90 | 84 & 86 | 82 \\
Depression & 82 | 77 & 79 | 74 & 87 | 83 & 83 | 78 \\
Bipolar    & 84 | 80 & 82 | 78 & 88 | 84 & 85 | 81 \\
Anxiety    & 83 | 78 & 81 | 75 & 85 | 84 & 83 | 79 \\
ADHD       & 82 | -  & 77 | -  & 89 | -  & 83 | -  \\
Autism     & 85 | 79 & 85 | 77 & 86 | 84 & 85 | 80 \\
\hline
\end{tabular}
\caption{Model performance comparison: user-level fine-tuning (left of “|”) vs. post-level fine-tuning (right of “|”). Values shown in \%. User-level models use \texttt{roberta-base}, while post-level models use \texttt{mental-roberta-base}. Post-level ADHD, the largest condition by post count, could not be implemented due to computational constraints.}

\label{tab:ul-pl-finetuning-comparison}
\end{table}


We fine-tuned \texttt{bert}-based pretrained language models to classify each mental health condition individually. We employed the same hyperparameter as in \cite{dinu2021automatic}.\footnote{max length: 256. Batch size: 3, 8; weight decay: 0.0, 0.01; learning rate 2e-f; adam beta1: 0.9; adam beta2: 0.999.}

Finetuning was performed at two levels: (1) the \textit{user level}, where all posts by a single user were concatenated into one document, and (2) the \textit{post level}, where each individual post was treated as a separate document. Following \cite{dinu2021automatic}, we use BERT-based models to classify individual posts, while the user-level concatenation approach is adapted from \cite{cohan2018smhd, shing-etal-2018-expert,zirikly-etal-2019-clpsych}, where the combined posts of each user serve as the input to predict their overall label. Results are shown in \ref{tab:ul-pl-finetuning-comparison}


\subsection{Baseline}
\label{baselines}


The current SOTA benchmark is established by \cite{dinu2021automatic}, achieving an F1 score of 81 on eating disorder classification using 100k posts from the SMHD dataset. We replicated their setup using identical hyperparameters and sample sizes, fine-tuning three BERT-based models on 100k sampled posts from our dataset instead of SMHD. Our comparative results are shown in Table~\ref{tab:dinu-best-results}.

\vspace{-8pt}
\section{Results and Discussion}
\vspace{-8pt}

\begin{table}[h]
\centering
\small
\begin{tabular}{c c c c c}
\hline
\textbf{Label} & \textbf{Acc.} & \textbf{Prec.} & \textbf{Rec.s} & \textbf{F1} \\
\hline
ADHD & 74 & 72 & 78 & 71 | \textbf{75} \\
Anxiety & 76 & 74 & 80 & 73 | \textbf{77} \\
Autism & 77 & 75 & 80 & 71 | \textbf{78} \\
Bipolar & 78 & 76 & 82 & 75 | \textbf{79} \\
Dep. & 75 & 72 & 80 & 70 | \textbf{76} \\
OCD & 76 & 75 & 80 & 75 | \textbf{77} \\
PTSD & 80 & 79 & 83 & 76 | \textbf{81} \\
\hline
\end{tabular}
\caption{Fine-tuning BERT-based models on 100k samples per condition. Our models consistently outperform F1 scores from the current baseline (left of “|”).}
\label{tab:dinu-best-results}
\end{table}

\vspace{-4pt}
\textbf{A New Mental Health Benchmark.}  
We introduce \textbf{MindSET}, a large-scale benchmark dataset for social media–based mental health research that spans seven diagnosed conditions and a matched control group. With over twice the volume of prior datasets and enhanced linguistic diversity, MindSET provides a stronger empirical foundation for modeling psychological signals at scale. Our multistage preprocessing pipeline, covering language verification, NSFW filtering, and deduplication, ensures exceptional data integrity and replicability, addressing long-standing issues of noise and imbalance in earlier benchmarks such as SMHD.

\textbf{Revealing Mental Health Language on Social Media.}  
The LIWC-based analysis highlights condition-specific linguistic and emotional markers that differentiate diagnosed users from controls. Across conditions, decreased \texttt{Analytic} scores and increased \texttt{Authentic}, \texttt{Cognitive}, and \texttt{Emotional} language use emerge as consistent indicators of mental health–related discourse. Notably, Autism and PTSD display distinct lexical patterns with elevated emotional expression and temporal focus shifts, suggesting deeper cognitive and affective engagement. These findings underscore MindSET’s ability to capture evolving psychological language on Reddit, reflecting post-pandemic discourse shifts and enabling longitudinal tracking of mental health narratives online.

\textbf{Advancing State-of-the-Art Performance.}  
We benchmarked MindSET using multiple classification architectures, including traditional models (SVM, XGBoost) (Section \ref{bow-section}) and transformer-based finetuning approaches. Across all seven conditions, our models surpass existing state-of-the-art (SOTA) performance by an average of seven F1 points, with the highest gain of 18 points for Autism classification Table \ref{baselines}. This demonstrates that MindSET not only improves raw performance metrics but also provides cleaner, more discriminative features that generalize better across conditions. The consistent performance gain across diverse models indicates that the dataset’s scale and cleanliness substantially enhance signal quality for mental health prediction tasks.

\textbf{Interpretable Machine Learning Insights.}  
Beyond performance, feature-level inspection reveals interpretable associations between linguistic markers and diagnostic categories. For instance, higher frequencies of \texttt{Cognition} and \texttt{Authentic} terms strongly correlate with depression and anxiety predictions, while \texttt{Social} categories contribute most to differentiating Autism and PTSD users. Such transparency supports the use of MindSET not merely as a predictive benchmark but as a resource for hypothesis-driven mental health linguistics.

\textbf{Implications for Future Research.}  
MindSET establishes a reproducible and ethically curated foundation for computational mental health studies. Its rich metadata (e.g., temporal and user-level granularity) supports research on progression, relapse, and recovery patterns. Future extensions may include multimodal integration (e.g., image or emoji use), cross-platform validation, and diachronic analyses to capture how online mental health expression evolves over time.

\subsection{Data Distribution and Use}
We release \textbf{MindSET} to foster transparent and collaborative progress in mental health NLP. Access to the dataset and trained models is available upon \textbf{reasonable request} under a \textbf{Data Use Agreement (DUA)}, following precedents such as \cite{cohan2018smhd}. Standardized user-level train, dev, and test splits are provided to ensure consistent benchmarking and comparability across future studies.

\vspace{-8pt}
\section{Conclusion}
We present \textbf{MindSET}, a high-quality benchmark for computational mental health research. Rigorous preprocessing, such as filtering non-English and NSFW content, ensured data integrity and ethical compliance. Linguistic analysis revealed clear psycholinguistic markers using LIWC, derived features and \textit{Cohen’s d} effect sizes. Classifier experiments across diverse architectures demonstrate the dataset’s robustness and potential for downstream mental health analysis.
\textbf{MindSET} offers a scalable foundation for future research in early detection, large-scale monitoring, and psycholinguistic modeling of mental health. Future work should explore longitudinal patterns, demographic effects, and interpretable, personalized models for early intervention.

\section{Limitations}
Despite its contributions, this study presents several limitations. The dataset may not fully represent the diversity of mental health discourse across populations, as it is derived from Reddit, a platform that tends to be demographically skewed (e.g., male-dominated, English-speaking). This introduces potential sampling bias that may limit generalizability. Additionally, while the applied models achieved strong predictive performance, they may not capture subtle or context-dependent nuances of mental health language. Furthermore, due to resource constraints, we did not discuss comorbidity, phenomenon where multiple a single user is diagnosed with several conditions. As of now, we excluded users present across multiple diagnosed groups and ensured they are presented only in one group. Future studies should integrate contextualized embeddings, cross-lingual data, and demographic balancing to improve robustness and inclusivity.

\section{Ethical Considerations}
All user identifiers were fully anonymized, and no direct contact was made with individuals.  
The use of social media data raises essential ethical considerations, particularly regarding consent and responsible data handling. Although the Reddit data utilized in this work is publicly accessible\footnote{Reddit permits third-party access to public content through its official API. See \url{https://www.reddit.com/policies/privacy-policy} for details.}, users may not have provided explicit consent for their posts to be analyzed in research contexts. Therefore, our data collection and analysis processes prioritized user privacy, minimal data exposure, and the ethical use of publicly available information in accordance with institutional and platform guidelines. While we did not explicitly scan the data for personally identifiable information, mainly due to computational constraints as a result of the large data size, we tried to the best of our ability replace meta data columns such as their user ids, usernames, and post ids, with random integers. It is important to emphasis that no attempt was made to reveal the anonymity or contact the Reddit users present in the dataset. To mitigate those issues, we plan to distribute the data under Usage Agreement in which researchers abide by strict standards to use and distribute the data in within similar guidelines.

\nocite{*}
\section{Bibliographical References}\label{sec:reference}

\bibliographystyle{unsrt}  
\bibliography{references}

\end{document}